\documentclass[11pt]{article}

\usepackage[final]{acl}

\usepackage{times}
\usepackage{latexsym}

\usepackage[T1]{fontenc}

\usepackage[utf8]{inputenc}

\usepackage{microtype}

\usepackage{inconsolata}

\usepackage{graphicx}

\usepackage{amsmath,amssymb,amsfonts}
\usepackage{algorithm}
\usepackage{algorithmic}
\usepackage{booktabs}      
\usepackage{multirow}
\usepackage[table]{xcolor} 
\usepackage{array}
\usepackage{makecell}
\usepackage{subcaption}
\title{Not All Preferences Deserve Gradients: \\ Understanding Gradient Utility in Offline Reasoning Alignment}

\author{
 \textbf{Hui Wu\textsuperscript{1}},
 \textbf{Hengyi Cai\textsuperscript{1}},
 \textbf{Jinman Zhao\textsuperscript{2}},
 \textbf{Xinran Chen\textsuperscript{1}}
\\
\textbf{Ziheng Li\textsuperscript{4}},
\textbf{Zhejun Zhao\textsuperscript{1}},
\textbf{Shuaiqiang Wang\textsuperscript{1}},
\textbf{Yuchen Li\textsuperscript{1,3}\thanks{Corresponding author}},
\textbf{Dawei Yin\textsuperscript{1}}
\\
 \textsuperscript{1}Baidu Inc. \textsuperscript{2}University of Toronto \textsuperscript{3}Shanghai Jiao Tong University
 \\
 \textsuperscript{4}School of Intelligence Science and Technology, Peking University
 \\
 \small{
    \texttt{wuhui21@mails.ucas.ac.cn,  yuchenli1230@gmail.com}
 }
}

\begin{document}
\maketitle
\begin{abstract}
Offline preference optimization aligns reasoning models from fixed chosen--rejected pairs, yet standard methods apply gradient updates from every pair regardless of its training value under the current policy.
We argue that this uniform treatment is wasteful and potentially harmful.
From the perspective of \emph{gradient utility}, we show that a pair's contribution depends jointly on informativeness and stability.
Pair utility drifts as the policy evolves, high-gradient samples can coincide with high-curvature regions, leading to noisy and destabilizing updates, and the most effective supervision comes from \emph{stable confident errors} where the model is reliably wrong yet curvature remains low.
These findings motivate \textbf{SAGE} (Stability-Aware Gradient Efficiency), which maintains difficulty-stratified candidate pools refreshed during training and selects pairs within each pool by a forward-pass signal-to-curvature score.
Only pairs with high current utility receive gradient computation; the rest are excluded from backpropagation.
On mathematical reasoning benchmarks across multiple model scales, SAGE outperforms full-data and size-matched baselines while producing substantially smoother optimization trajectories.
\end{abstract}

\section{Introduction}
Large reasoning models generate explicit chains of thought that can be verified and used to construct pairwise preferences for alignment~\cite{guo2025deepseek,jaech2024openai,li2025towards}.
Direct preference optimization (DPO) and its variants have become a standard choice in this setting: they avoid reward modeling and online rollout generation, and instead optimize the policy directly on fixed chosen--rejected pairs~\cite{rafailov2023direct,meng2024simpo,ethayarajh2024model}.
However, standard offline preference optimization does not differentiate among preference pairs once they enter the dataset.
Each pair is treated as eligible for backpropagation regardless of its utility under the current policy.
This uniform use of supervision can be inefficient and even detrimental for long-chain-of-thought (Long-CoT) reasoning, where noisy or unstable updates may disrupt the multi-step reasoning trajectories that determine final-answer accuracy~\cite{xiong2024search}.

\begin{figure}
    \centering
    \includegraphics[width=1\linewidth]{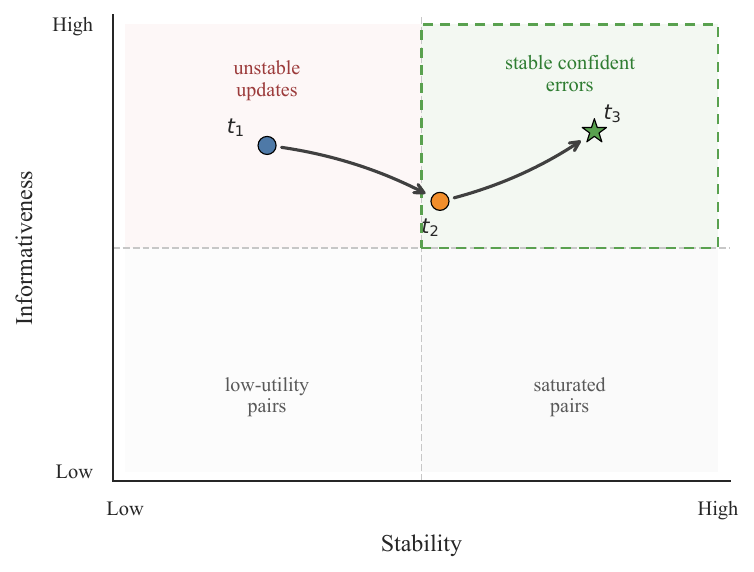}
    \caption{Schematic illustration of preference-pair utility in the informativeness--stability space. As the policy evolves, the utility of a pair can shift across training stages. SAGE selects pairs that combine strong corrective signal with stable local optimization behavior.}
    \label{fig1}
\end{figure}

As the policy evolves during training, some pairs become saturated and yield negligible gradients, while others lie in high-curvature regions of the local loss landscape where small perturbations can amplify noisy updates.
The central question, then, is not merely which loss to optimize, but which preference pairs should contribute gradient updates at each training stage.
We formalize this question through \textbf{gradient utility}, a pair-level measure of training value that separates two factors: \emph{informativeness}, the strength of the corrective signal, and \emph{stability}, the reliability of the resulting update under local curvature.

Through this lens, we conduct a systematic diagnosis of Long-CoT preference optimization and identify three findings.
First, gradient utility is policy-dependent: pair rankings shift substantially across training checkpoints, making static selection poorly matched to the evolving policy (\S\ref{sec:finding_dynamic}).
Second, large gradients do not imply high utility; they often arise in high-curvature regions and induce noisy, destabilizing updates rather than useful corrections (\S\ref{sec:finding_gradient}).
Third, the most effective pairs are \emph{stable confident errors}: cases that combine strong corrective signal with well-conditioned local geometry (\S\ref{sec:finding_confident}).
Together, these findings reveal a blind spot in existing offline preference methods: they optimize over all available pairs without explicitly assessing whether each pair provides a useful gradient for the current policy.

Motivated by this analysis, we propose \textbf{SAGE} (Stability-Aware Gradient Efficiency), a policy-aware preference selection framework for offline alignment.
SAGE combines periodically refreshed, difficulty-stratified candidate pools with fine-grained utility-based selection.
The pool mechanism tracks the evolving policy without repeatedly rescoring the full corpus, while the utility score ranks pairs using a Newton-inspired signal-to-curvature criterion.
Together, these components prioritize stable confident errors and skip saturated or high-curvature pairs for the current training stage.
Unlike online RL methods such as GRPO~\cite{shao2024deepseekmath}, which improve policies through new response generation and reward-driven exploration, SAGE addresses a complementary offline question: how to allocate gradient computation over an already-collected preference corpus.

Our contributions are threefold.
First, we introduce \emph{gradient utility} as a diagnostic framework for offline preference optimization in Long-CoT reasoning.
Second, we show empirically that pair utility is policy-dependent, that large gradients can be destabilizing under high curvature, and that stable confident errors provide the most effective supervision.
Third, we propose \textbf{SAGE}, a gradient-utility-aware selection method that improves optimization stability and reasoning accuracy across multiple model scales while reducing unnecessary gradient computation.
\section{Preliminaries}
\paragraph{Notation.}
Let $x$ denote a prompt and $y$ a response.
We define $\pi_\theta(y \mid x)$ for the trainable policy and $\mu(y \mid x)$ for a fixed reference policy.
Preference supervision consists of triples $(x, y^w, y^l)$ where $y^w \succ y^l$.
In mathematical reasoning, $x$ is a problem statement, $y^w$ a correct solution, and $y^l$ an incorrect one.
The implicit reward is defined as $r_\theta(x, y) = \beta \log \frac{\pi_\theta(y \mid x)}{\mu(y \mid x)}$ with temperature $\beta > 0$.

\paragraph{Noise-Contrastive Alignment (NCA).}
We adopt NCA~\cite{chen2024noise} as the preference objective throughout this work.
Given a prompt $x$ with $K$ candidate responses $\{y_i\}_{i=1}^K$, NCA computes normalized weights $w_i(x)=\frac{\exp(r_i/\alpha)}{\sum_{j=1}^K\exp(r_j/\alpha)}$ and optimizes:
\begin{equation}
\mathcal{L}_{\mathrm{NCA}}
= -\sum_{i=1}^K\!\left[
w_i(x)\,\log \sigma(r_i)
+\frac{1}{K}\,\log \sigma(-r_i)
\right]
\label{eq:nca_loss}
\end{equation}
where $r_i = r_\theta(x, y_i)$ and $\alpha > 0$ is a temperature.
Compared to DPO~\cite{rafailov2023direct}, NCA encourages high-reward responses while regularizing uniformly weighted negatives, improving training stability.
Structurally, the NCA loss decomposes into per-response binary classification terms $\ell(r) = -\log\sigma(zr)$ with $z \in \{+1, -1\}$, which allows us to analyze the optimization dynamics of each response independently.
This decomposition is central to the gradient utility analysis in Section~\ref{sec:what_is_gu} and the derivation of the SAGE score in Section~\ref{section3.3}.
\section{Gradient Utility: Diagnosis and Findings}
\label{sec:gradient_utility}

We develop \emph{gradient utility} as a diagnostic framework for understanding when preference pairs yield useful gradients in Long-CoT reasoning alignment.
This section defines gradient utility through a local signal-to-curvature view of pairwise preference optimization, then presents three empirical findings that motivate the design of SAGE.
Unless otherwise specified, all analyses use Qwen2.5-1.5B-Instruct trained on our preference corpus (Section~\ref{section5.1}).

\subsection{Defining Gradient Utility}
\label{sec:what_is_gu}

Consider a preference pair $(x,y^w,y^l)$ under the current policy $\pi_{\theta_t}$.
Let $m_{\theta_t}(x,y^w,y^l)$ denote the signed pairwise preference margin, where larger values indicate stronger preference for $y^w$ over $y^l$.
For DPO-style objectives, this margin corresponds to the implicit reward difference between the two responses.
Locally, we write the pairwise loss as
\begin{equation}
    \ell(m) = -\log \sigma(m)
    \qquad
    p = \sigma(m)
\end{equation}
where $p$ is the current confidence that the pair is ordered correctly.

The margin-level gradient and curvature are
\begin{equation}
    \left|\frac{\partial \ell}{\partial m}\right| = 1-p
    \qquad
    \frac{\partial^2 \ell}{\partial m^2} = p(1-p)
\end{equation}
We then define gradient utility as
\begin{equation}
    \mathrm{GU}(x,y^w,y^l;\theta_t)
    =
    \frac{g^2}{c+\epsilon}
    =
    \frac{(1-p)^2}{p(1-p)+\epsilon}
    \label{eq:gradient_utility}
\end{equation}
where $g=1-p$ is the informativeness (corrective signal strength), $c=p(1-p)$ is the local curvature, and $\epsilon>0$ smooths the denominator.
This score is a margin-space signal-to-curvature proxy: it favors pairs with strong corrective signal and low curvature.
It is largest when the model is confidently wrong, moderate near uncertain predictions, and vanishes when the model is already correct.
We call high-utility pairs \emph{stable confident errors}: the model is reliably wrong yet the local geometry permits a stable correction.

\subsection{Finding 1: Pair Utility Drifts with the Policy}
\label{sec:finding_dynamic}

\begin{figure}[t]
    \centering
    \includegraphics[width=\linewidth]{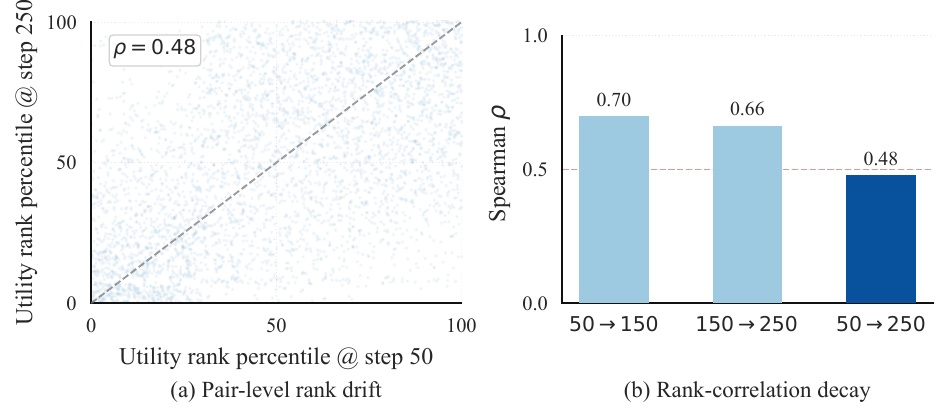}
    \caption{Pair-level gradient utility rankings shift substantially as the policy evolves.}
    \label{fig2}
\end{figure}

\paragraph{Observation.}
We compute gradient utility scores for all 4{,}134 pairs at three training checkpoints: steps 50, 150, and 250.
Figure~\ref{fig2} reports the Spearman rank correlation of pair-level utility rankings across checkpoints.
The correlation between early and late checkpoints drops below 0.5, indicating that the relative utility ordering is substantially reshuffled as the policy evolves.

\paragraph{Analysis.}
This drift follows from the policy-dependence of the confidence term $p=\sigma(m_{\theta_t})$.
As training changes $\theta_t$, the same pair may move from informative to saturated, or from unstable to locally well conditioned.
The utility ordering is therefore not a static property of the dataset but a dynamic function of the current policy.
A selection strategy that fixes a subset before training cannot adapt to this redistribution, motivating SAGE's refreshable pool design that periodically re-evaluates pair utility under the evolving policy.

\subsection{Finding 2: Large Gradients Are Often Unstable}
\label{sec:finding_gradient}

\begin{figure}[t]
    \centering
    \includegraphics[width=\linewidth]{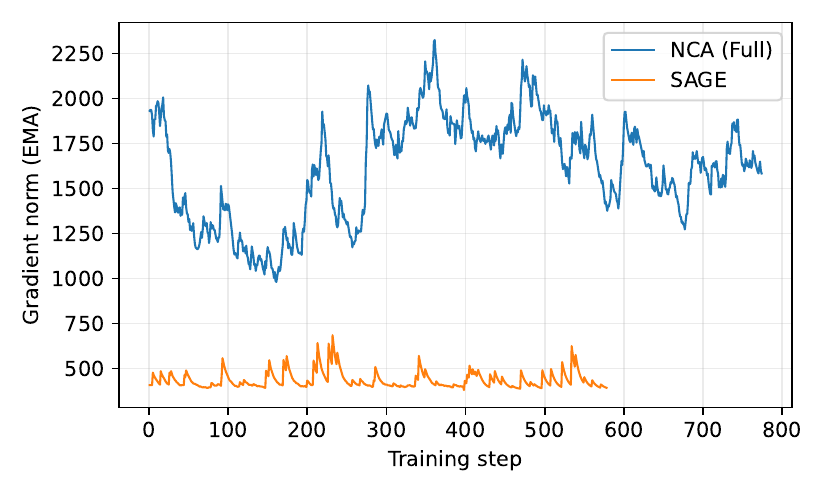}
    \caption{Gradient-norm dynamics during training. Full-data training exhibits frequent spikes; gradient-utility-based selection produces smoother optimization.}
    \label{fig3}
\end{figure}

\paragraph{Observation.}
We track gradient-norm dynamics throughout training under two conditions: standard training on the full preference corpus, and training on gradient-utility-selected subsets with the same effective data budget.
Figure~\ref{fig3} shows that full-data training exhibits frequent, large gradient spikes throughout optimization.
These spikes are consistent with unstable high-curvature regions near the decision boundary ($p \approx 0.5$), where local updates are more sensitive to perturbations.
By contrast, training on gradient-utility-selected data, which discounts high-curvature pairs, produces substantially smoother trajectories with a lower peak-to-mean ratio.

\paragraph{Analysis.}
For a pairwise loss, the parameter-space gradient decomposes as
\begin{equation}
    \nabla_\theta \ell(m_\theta)
    =
    -(1-p)\,\nabla_\theta m_\theta
\end{equation}
A large gradient norm may reflect a genuinely informative error, or it may arise from a large $\|\nabla_\theta m_\theta\|$ in a high-curvature region.
We stress that Figure~\ref{fig3} plots the \emph{parameter-space} norm $\|\nabla_\theta \ell\| = (1-p)\|\nabla_\theta m_\theta\|$, which is not determined by the margin-level quantities alone: in margin space $|\partial \ell / \partial m|$ is maximized as $p \to 0$, whereas the curvature $p(1-p)$ peaks at $p \approx 0.5$.
The spikes in Figure~\ref{fig3} are therefore suggestive but cannot, by themselves, be attributed to high curvature.
To establish the link directly, we measure the local curvature of each pair along its own gradient direction using Hessian-vector products and compare the two groups that SAGE separates (protocol in Appendix~\ref{appendix:curvature}): the high-curvature pairs that SAGE discounts produce gradient spikes (norm exceeding twice the running median) in $21\%$ of the steps in which they participate, versus $5\%$ for the low-curvature pairs that SAGE retains.
Gradient utility explicitly balances corrective signal against local curvature, discounting pairs whose large gradients are likely to produce noisy updates.
The smoother optimization in Figure~\ref{fig3} supports this interpretation: filtering high-curvature pairs removes destabilizing spikes while preserving useful corrective signal.
The ablation in Section~\ref{section5.2} provides causal confirmation, removing curvature discounting (w/o $c$) reintroduces gradient spikes and degrades the harder benchmarks most severely (Minerva $-2.90$, Gaokao $-2.90$), consistent with the mechanism identified here.
This distinction is especially important in Long-CoT reasoning, where unstable updates can disrupt multi-step reasoning trajectories.

\begin{figure*}[t]
    \centering
    \includegraphics[width=\linewidth]{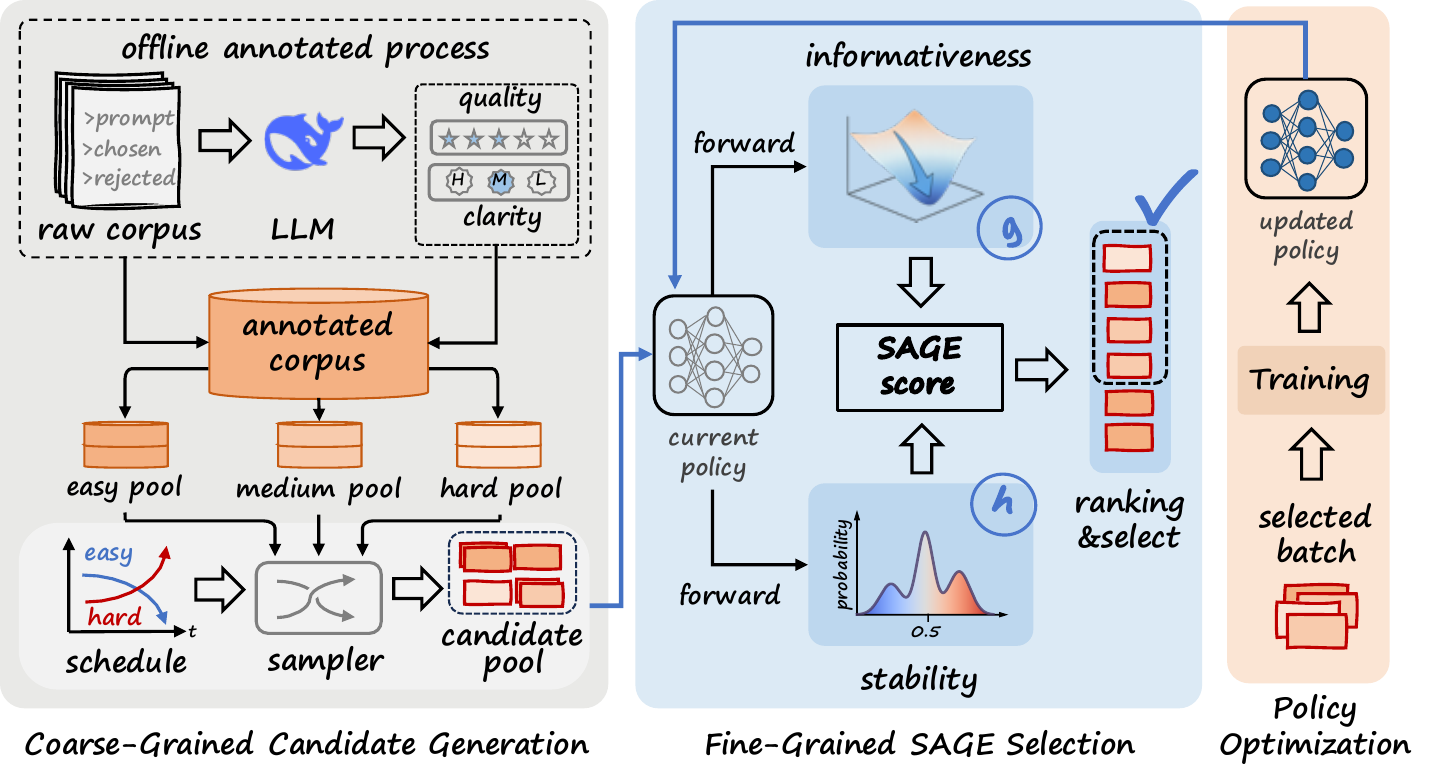}
    \caption{Overall framework of SAGE. A candidate pool is constructed from difficulty-stratified subsets at each interval, then pairs are scored by gradient utility and only high-scoring pairs receive gradient computation.}
    \label{fig4}
\end{figure*}

\subsection{Finding 3: Stable Confident Errors Are Most Effective}
\label{sec:finding_confident}

\paragraph{Synthesis.}
Findings~1 and 2 jointly identify the regime of maximal gradient utility.
A useful pair must satisfy two conditions simultaneously: high informativeness ($g$ large, so the model is confidently wrong and the corrective signal is strong) and low curvature ($c$ small, so the update is locally stable).
Pairs satisfying both conditions, which we term \emph{stable confident errors}, correspond to the high-informativeness, low-curvature regime.

This characterization excludes three other regimes:
(i)~saturated pairs (low $g$, low $c$), where the model already assigns the correct preference and provides negligible signal;
(ii)~informative but unstable pairs (high $g$, high $c$), where strong signal exists but high curvature makes the update unreliable;
and (iii)~weak and noisy pairs (low $g$, high $c$), which are neither informative nor stable.

\paragraph{Verification.}
The ablation in Section~\ref{section5.2} confirms this prediction: removing $c$ (retaining only $g$) or removing $g$ (retaining only $c$) both degrade performance, confirming that neither informativeness nor stability alone suffices.
The optimal criterion is pairs that maximize corrective signal relative to local curvature, exactly the stable confident errors that SAGE targets.

\section{Method}
\label{section4}

The diagnostic findings in Section~\ref{sec:gradient_utility} establish that effective preference selection requires both policy-adaptivity (Finding~1) and joint consideration of informativeness and stability (Findings~2--3).
We now describe \textbf{SAGE} (Stability-Aware Gradient Efficiency), which operationalizes these principles through two mechanisms: coarse-grained candidate pools that enable efficient on-policy re-scoring (Section~\ref{section3.2}), and a fine-grained utility score that identifies stable confident errors within each pool (Section~\ref{section3.3}).
Figure~\ref{fig4} illustrates the overall pipeline.

\subsection{Problem Setup}\label{section3.1}
We consider the alignment of a reasoning policy $\pi_\theta$ using a static dataset $\mathcal{D} = \{(x_i, y_i^w, y_i^l)\}_{i=1}^N$, where $x_i$ is a prompt and $(y_i^w, y_i^l)$ are the preferred and rejected responses (see Appendix~\ref{Data Construction} for data construction details).
The standard approach minimizes a loss $\mathcal{L}(\theta)$ over the full data expectation.
However, as Finding~1 demonstrates, minimizing the average loss over a static distribution ignores the dynamic nature of pair utility.
Our goal is to dynamically select a sub-batch $\mathcal{B}_t \subset \mathcal{D}$ at each training interval that maximizes gradient utility, prioritizing pairs that are both informative and stable under the current policy $\pi_{\theta_t}$.

\paragraph{On-Policy Selection.}
We clarify an important terminological distinction.
In this work, \emph{on-policy selection} refers to evaluating the utility of existing static preference pairs using the current model state $\pi_{\theta_t}$, rather than generating new responses through online exploration.
This allows SAGE to maintain policy-adaptivity without the computational cost of online rollout generation.

\subsection{Coarse-Grained Candidate Generation}\label{section3.2}
Evaluating gradient utility for all $N$ samples at every training step is computationally prohibitive.
SAGE addresses this through a refreshable pool strategy that constructs candidate pools from difficulty-stratified subsets, providing a practical approximation to full on-policy re-scoring.

\paragraph{Difficulty-Based Partitioning.}
We partition $\mathcal{D}$ into three subsets $\{\mathcal{D}_{\text{easy}}, \mathcal{D}_{\text{medium}}, \mathcal{D}_{\text{hard}}\}$ based on off-policy difficulty priors.
Each preference pair is annotated with two auxiliary judgments via an LLM-as-judge:
(i) a \emph{preference clarity} label $c_i \in \{\text{high}, \text{medium}, \text{low}\}$, indicating how clearly the chosen response is preferred over the rejected one, and
(ii) a quality score $q_i \in \{1,\dots,5\}$ for the rejected response (details in Appendix~\ref{Preference Annotation}).
Based on these annotations, we bucket samples into three difficulty strata:
\begin{align}
\mathcal{D}_{\text{easy}} &= \{ i \mid c_i = \text{high},~ q_i \le 2 \} \notag \\
\mathcal{D}_{\text{medium}} &= \{ i \mid c_i = \text{medium},~ q_i \in \{2,3\} \} \notag \\
\mathcal{D}_{\text{hard}} &= \mathcal{D} \setminus (\mathcal{D}_{\text{easy}} \cup \mathcal{D}_{\text{medium}})
\end{align}
These static priors serve only as a coarse stratification for pool construction; the fine-grained selection within each pool is performed on-policy.

\paragraph{Dynamic Pool Construction.}
Training is divided into $K$ intervals, each associated with a refreshable candidate pool $\mathcal{P}_k$.
We partition $\mathcal{D}$ into $K$ non-overlapping pools:
\begin{equation}
    \mathcal{D} = \bigcup_{k=1}^{K} \mathcal{P}_k, \quad \text{s.t.} \quad \mathcal{P}_i \cap \mathcal{P}_j = \varnothing
\end{equation}
where each pool contains $M = |\mathcal{D}|/K$ samples.
The composition of $\mathcal{P}_k$ is not uniform; it is constructed by sampling from difficulty subsets according to a time-dependent mixing ratio $\rho(k)$:
\begin{equation}
   \rho_c(k) = \rho_c^{\text{start}} + \left(\rho_c^{\text{end}} - \rho_c^{\text{start}}\right) \cdot \frac{k}{K}
   \label{eq:schedule}
\end{equation}
where $c \in \{\text{easy}, \text{medium}, \text{hard}\}$ and $\sum_c \rho_c(k) = 1$.
This schedule implements a coarse curriculum: early pools contain more easy pairs that provide stable initial supervision, while later pools shift toward harder pairs as the policy matures.
This directly addresses Finding~1 by ensuring that the candidate distribution evolves with the policy, without requiring full-dataset re-evaluation at every step.

\subsection{Fine-Grained Selection via the SAGE Score}\label{section3.3}
Within each candidate pool $\mathcal{P}_k$, we apply a fine-grained filtering mechanism based on gradient utility.
The key insight from Section~\ref{sec:what_is_gu} is that useful pairs maximize the ratio of gradient signal to local curvature.
We now derive a tractable scoring function that operationalizes this principle.

\paragraph{From Newton Decrement to SAGE Score.}
The Newton decrement $\lambda(x)$ provides a classical measure of effective optimization progress:
\begin{equation}
\frac{1}{2}\lambda^2(x) \approx \frac{1}{2}
\nabla_\theta \mathcal{L}(x)^\top \mathbf{H}^{-1}(x)\nabla_\theta \mathcal{L}(x)
\end{equation}
where $\lambda^2(x)/2$ estimates the expected loss reduction under a Newton step.
Computing the full Hessian inverse $\mathbf{H}^{-1}$ is infeasible for large-scale models.
Instead, we exploit the per-response decomposition of the NCA loss to derive a lightweight, sample-level approximation.

Since the NCA loss decomposes into binary classification terms $\ell(r) = -\log\sigma(zr)$, we can analyze each response independently.
For a single response with confidence $p = \sigma(zr)$:
\begin{itemize}
    \item Squared gradient magnitude: $g^2 = (\partial_r \ell)^2 = (1-p)^2$
    \item Local curvature: $c = \partial_r^2 \ell = p(1-p)$
\end{itemize}
The per-response Newton decrement contribution is then $g^2/c$, which is exactly the gradient utility ratio from Eq.~\ref{eq:gradient_utility}.

Reasoning trajectories vary significantly in length, which can bias the score toward longer responses.
To mitigate this, we introduce length normalization by the total response token count $L_i$.
The SAGE score for pair $i$ is:
\begin{equation}
\begin{aligned}
s_i(x) &= \frac{1}{L_i} \sum_{z \in \{y^w, y^l\}}
\frac{g_z^2}{c_z + \epsilon} \\
&= \frac{1}{L_i} \sum_{z \in \{y^w, y^l\}}
\frac{(1-p_z)^2}{p_z(1-p_z) + \epsilon}
\end{aligned}
\label{eq:sage_score}
\end{equation}
where $\epsilon$ is a Tikhonov damping term for numerical stability in near-deterministic regimes.
This score requires only a forward pass to compute $p_z$ for each response; no gradient computation or Hessian estimation is needed.
\paragraph{Dynamic Batch Construction.}
After scoring all candidates in $\mathcal{P}_k$, SAGE performs hard filtering.
Samples are ranked in descending order of their SAGE scores, and only the top-$\gamma_k$ fraction is retained:
\begin{equation}
\mathcal{B}_k =
\left\{ x_i \in \mathcal{P}_k \mid
\text{rank}(s_i) \le \gamma_k \cdot |\mathcal{P}_k| \right\}
\end{equation}
where $\gamma_k$ is a time-dependent selection ratio.
The policy is updated exclusively on $\mathcal{B}_k$; low-scoring samples are excluded from backpropagation.
This implements the gradient utility principle: only pairs that are simultaneously informative and stable contribute to parameter updates.

\paragraph{Connection to Findings.}
The SAGE score directly operationalizes the three findings:
it adapts to the current policy through on-policy forward-pass scoring (Finding~1),
penalizes high-curvature samples via the denominator $c + \epsilon$ (Finding~2),
and prioritizes stable confident errors where $g^2/c$ is maximized (Finding~3).
The hard truncation further ensures that weak-signal pairs ($g \to 0$) and maximally uncertain pairs ($c \gg 0$) are physically removed from the training batch.
\section{Experiments}
\subsection{Experimental Setup}\label{section5.1}

\paragraph{Models and Baselines.}
We use the Qwen2.5-Instruct family~\cite{qwen2025qwen25technicalreport} at three scales (1.5B, 3B, 7B) and additionally evaluate on Qwen3-4B-Instruct-2507.
For 1.5B and 3B, we use the base Instruct checkpoints (see Appendix~\ref{sft_results}); for 7B, we use the post-SFT checkpoint.
Baselines include NCA on the full corpus, NCA on a size-matched random subset, alternative objectives (SimPO~\cite{meng2024simpo}, KTO~\cite{ethayarajh2024model}), and data selection methods: static gradient-norm ranking, log-probability filtering, dynamic pair-loss selection (top-$\gamma_k$ pairs by current NCA loss at each interval), and gradient-similarity selection adapted from LESS~\cite{xia2024less} to pair-level preference losses.

\paragraph{Training Data.}
We leverage the Light-R1 datasets~\cite{wen2025light}.
SFT warm-up uses 76k samples from Stage~1.
For preference optimization, we construct \textbf{4,134} pairs by deduplicating the original Light-R1 preference set and generating additional rejected responses (Appendix~\ref{Data Construction}).

\begin{table*}[t]
\centering
\small
\setlength{\tabcolsep}{4.2pt} 
\begin{tabular}{l *{8}{c} | c}
\toprule
\textbf{Method} & \textbf{GSM8K} & \textbf{M500} & \textbf{Minerva} & \textbf{Gaokao} & \textbf{Olymp.} & \textbf{Coll.} & \textbf{AIME24} & \textbf{AMC23} & \textbf{Avg.} \\
\midrule

\rowcolor[gray]{0.95} \multicolumn{10}{c}{\textbf{Qwen2.5-1.5B-Instruct}} \\
Vanilla & 73.70 & 54.60 & 16.90 & 46.20 & \textbf{22.70} & \textbf{38.40} & 6.70 & 25.00 & 35.53 \\
w/ NCA (Full) & 74.70 & 56.20 & 19.50 & 47.30 & 20.00 & 38.00 & \textbf{10.00} & 22.50 & 36.03 \\
w/ NCA (Random) & 73.50 & 56.40 & 19.10 & 48.60 & 19.60 & 37.90 & 3.30 & 25.00 & 35.43 \\
\textbf{SAGE (Ours)} & \textbf{74.80} & \textbf{57.20} & \textbf{20.20} & \textbf{50.40} & 21.50 & 38.10 & \textbf{10.00} & \textbf{27.50} & \textbf{37.46} \\

\cmidrule(lr){1-10}
\multicolumn{10}{l}{\textit{\footnotesize \color{gray} Alternative preference objectives (on 1.5B)}} \\
\hspace{1em} + SimPO & 74.50 & 56.20 & 19.10 & 48.30 & 21.80 & 37.90 & \textbf{10.00} & 25.00 & 36.60 \\
\hspace{1em} + KTO & 74.10 & 55.60 & 19.50 & 49.10 & 19.40 & 38.10 & 6.70 & 22.50 & 35.63 \\

\addlinespace[2pt]
\multicolumn{10}{l}{\textit{\footnotesize \color{gray} Data selection baselines (on 1.5B)}} \\
\hspace{1em} + Gradient-based & 73.80 & 56.80 & 19.50 & 47.80 & 20.90 & \textbf{38.40} & 3.30 & 25.00 & 35.69 \\
\hspace{1em} + Log-P & 73.50 & 54.80 & 18.00 & 48.10 & 21.30 & 38.10 & 3.30 & 20.00 & 34.64 \\
\hspace{1em} + Dynamic Pair-Loss & 74.50 & 56.80 & 19.80 & 48.80 & 20.80 & 38.30 & 6.70 & 25.00 & 36.34 \\
\hspace{1em} + Grad-Similarity & 74.60 & 57.00 & 20.20 & 49.50 & 21.30 & 38.30 & 6.70 & 25.00 & 36.58 \\

\midrule
\rowcolor[gray]{0.95} \multicolumn{10}{c}{\textbf{Qwen2.5-3B-Instruct}} \\
Vanilla & 86.90 & 65.20 & 25.70 & 56.40 & \textbf{27.70} & 44.50 & 6.70 & 47.50 & 45.08 \\
w/ NCA (Full) & 86.40 & 65.60 & 27.20 & 56.90 & 27.00 & 44.90 & 10.00 & 50.00 & 46.00 \\
w/ NCA (Random) & 87.00 & 65.20 & 26.10 & 56.40 & 26.50 & 45.00 & 0.00 & 45.00 & 43.90 \\
\textbf{SAGE (Ours)} & \textbf{87.50} & \textbf{66.00} & \textbf{28.30} & \textbf{58.23} & \textbf{27.70} & \textbf{45.14} & \textbf{13.30} & \textbf{55.00} & \textbf{47.65} \\

\midrule
\rowcolor[gray]{0.95} \multicolumn{10}{c}{\textbf{Qwen2.5-7B-Instruct}} \\
Vanilla & 92.30 & 81.60 & 28.30 & 69.90 & 45.30 & 42.40 & 23.30 & 57.50 & 55.08 \\
w/ NCA (Full) & 92.70 & 82.00 & 29.40 & 70.60 & \textbf{46.50} & 42.70 & 26.70 & 62.50 & 56.64 \\
w/ NCA (Random) & 91.30 & 79.40 & 26.80 & \textbf{71.40} & 43.00 & 42.70 & 20.00 & 57.50 & 54.01 \\
\textbf{SAGE (Ours)} & \textbf{93.10} & \textbf{82.80} & \textbf{33.10} & \textbf{71.40} & 45.50 & \textbf{43.10} & \textbf{33.30} & \textbf{70.00} & \textbf{59.04} \\

\midrule
\rowcolor[gray]{0.95} \multicolumn{10}{c}{\textbf{Qwen3-4B-Instruct-2507}} \\
Vanilla & 94.10 & 93.40 & 40.80 & 81.30 & 61.60 & \textbf{39.80} & 60.00 & 87.50 & 69.82 \\
w/ NCA (Full) & 94.00 & 93.80 & 41.50 & 81.80 & \textbf{63.30} & 39.70 & 60.00 & 92.50 & 70.83 \\
w/ NCA (Random) & 93.80 & 92.90 & 41.20 & 82.10 & 61.40 & 39.30 & 56.70 & 92.50 & 69.99 \\
\textbf{SAGE (Ours)} & \textbf{94.50} & \textbf{94.60} & \textbf{42.60} & \textbf{82.40} & 62.20 & \textbf{39.80} & \textbf{63.30} & \textbf{95.00} & \textbf{71.80} \\
\bottomrule
\end{tabular}
\caption{Main results on mathematical reasoning benchmarks across multiple model scales. \textbf{M500}: MATH500. \textbf{Olymp.}: OlympiadBench. \textbf{Coll.}: CollegeMath. Best results are bolded.}
\label{table1}
\end{table*}

\paragraph{Evaluation.}
We evaluate on GSM8K~\citep{cobbe2021training}, MATH500, Minerva-MATH~\citep{lewkowycz2022solving}, Gaokao23-Math~\citep{liao2024mario}, OlympiadBench~\citep{he2024olympiadbench}, CollegeMath~\citep{tang2024mathscale}, AMC23, and AIME24.
We report pass@1 on all benchmarks except AIME24 and AMC23, where we conduct 8 independent evaluation runs and report the average pass@8 to reduce variance from their small test sizes (30 and 40 problems, respectively); consistency of gains across four independent model scales further mitigates single-run noise.
Maximum generation length is 10{,}000 tokens.
Evaluation uses the SimpleRL-Reason framework.

\paragraph{Implementation.}
We build on 360-LLaMA-Factory with sequence parallelism.
Default settings: $\beta=0.1$, 3 epochs, batch size 16, 8$\times$A800 GPUs (sp=8), max sequence length 32{,}768.

\subsection{Main Results}
Table~\ref{table1} reports mathematical reasoning performance across multiple model scales.
``NCA (Full)'' trains on the full preference dataset; ``NCA (Random)'' uses a size-matched random subset for fair comparison with selective methods.

Across all scales, SAGE consistently improves over standard NCA in average accuracy, with gains particularly pronounced on harder benchmarks (e.g., +3.7 on Minerva and +7.5 on AMC23 at 7B) where multi-step reasoning is more susceptible to gradient instability (Section~\ref{sec:finding_gradient}).
Compared with alternative objectives (SimPO, KTO) and data selection heuristics, SAGE performs best on average, confirming that jointly considering informativeness and stability outperforms methods addressing only one dimension.
Among dynamic selection methods, both Dynamic Pair-Loss and Grad-Similarity outperform static heuristics, yet neither matches SAGE: loss-based selection fails to improve on competition-level benchmarks (AIME24, AMC23), confirming that high loss does not imply high utility when curvature is uncontrolled; gradient-similarity narrows the gap but still lacks explicit stability modeling.
SAGE achieves these gains while using fewer effective training tokens than NCA (Full), and scale-consistent behavior suggests gradient utility is a general property of preference optimization.
Table~\ref{table1} reports single-run results; Appendix~\ref{appendix:seed_variance} additionally reports a five-seed study on 1.5B with mean$\pm$std on all eight benchmarks, where SAGE improves the average accuracy over NCA (Full) ($37.3 \pm 0.6$ vs.\ $36.1 \pm 0.6$; paired $t$-test, $p \approx 0.008$).
As expected, the two small competition sets (AIME24, AMC23) exhibit the largest seed variance and overlapping per-seed intervals, so we base our conclusions on the aggregate rather than on these individual benchmarks.

\subsection{Ablation Study}\label{section5.2}

\begin{table}[h]
\centering
\renewcommand{\arraystretch}{1.2}
\resizebox{\linewidth}{!}{
\begin{tabular}{l cccc}
\toprule
\textbf{Setting} & \textbf{GSM8K} & \textbf{MATH500} & \textbf{Minerva} & \textbf{Gaokao} \\
\midrule
\textbf{SAGE} & \textbf{74.80} & \textbf{57.20} & \textbf{20.20} & \textbf{50.40} \\
\quad w/o $s_1$ & 74.50 {\color{red}(-0.30)} & 56.60 {\color{red}(-0.60)} & 19.60 {\color{red}(-0.60)} & 48.90 {\color{red}(-1.50)} \\
\quad w/o $s_2$   & 73.90 {\color{red}(-0.90)} & 55.30 {\color{red}(-1.90)} & 17.50 {\color{red}(-2.70)} & 47.10 {\color{red}(-3.30)} \\
\quad w/o $g$         & 74.00 {\color{red}(-0.80)} & 56.00 {\color{red}(-1.20)} & 18.00 {\color{red}(-2.20)} & 48.10 {\color{red}(-2.30)} \\
\quad w/o $c$        & 73.80 {\color{red}(-1.00)} & 55.40 {\color{red}(-1.80)} & 17.30 {\color{red}(-2.90)} & 47.50 {\color{red}(-2.90)} \\
\bottomrule
\end{tabular}
}
\caption{Ablation of SAGE components on Qwen2.5-1.5B-Instruct. $s_1$: coarse-grained pool construction; $s_2$: fine-grained selection; $g$: gradient signal; $c$: curvature discounting.}
\label{table2}
\end{table}

\noindent Table~\ref{table2} ablates SAGE's key components on Qwen2.5-1.5B-Instruct.
Removing curvature discounting (w/o $c$) produces the largest drops on harder benchmarks (Minerva $-2.90$, Gaokao $-2.90$), directly validating Finding~2: discounting high-curvature pairs is essential for stable optimization.
Removing the gradient signal term (w/o $g$) also degrades performance, confirming Finding~3: informativeness alone is insufficient without stability, and vice versa.
Removing coarse-grained pool construction ($s_1$) or fine-grained selection ($s_2$) causes further degradation, showing that both the policy-adaptive pool mechanism (motivated by Finding~1) and the utility-based scoring contribute independently.
Together, these results validate the two-dimensional gradient utility framework: both $g$ and $c$ are necessary, and neither coarse nor fine-grained selection alone suffices.

\subsection{Analysis}
\begin{table}[h]
\centering
\footnotesize
\setlength{\tabcolsep}{3pt}
\begin{tabular}{l cccc}
\toprule
\textbf{Method} & \textbf{ARC-C} & \textbf{GPQA-D} & \textbf{MMLU-Pro} & \textbf{Avg.} \\
\midrule
Vanilla & 63.70 & 25.30 & 24.70 & 37.90 \\
w/ NCA (Full) & 65.30 & 26.80 & \textbf{26.90} & 39.67 \\
w/ NCA (Random) & 64.30 & 23.70 & 25.00 & 37.67 \\
\midrule
\textbf{SAGE (Ours)} & \textbf{66.10} & \textbf{27.50} & 26.50 & \textbf{40.03} \\
\bottomrule
\end{tabular}
\caption{Performance on general-domain benchmarks beyond mathematical reasoning.}
\label{table6}
\end{table}

\paragraph{General Capability.}
Table~\ref{table6} evaluates the aligned 1.5B model on general-domain benchmarks.
SAGE achieves the best average, indicating that removing noisy high-curvature pairs preserves supervision diversity.

\begin{figure}
    \centering
    \includegraphics[width=\linewidth]{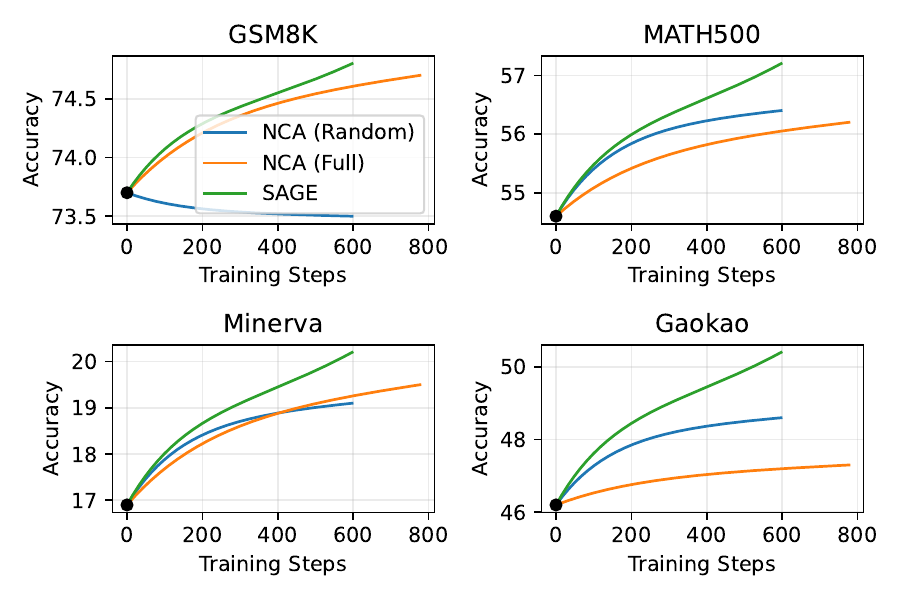}
    \caption{Evaluation performance over training steps.}
    \label{fig5}
\end{figure}

\paragraph{Training Dynamics.}
Figure~\ref{fig5} tracks evaluation performance over training steps.
Both NCA baselines exhibit early saturation, consistent with the utility drift phenomenon (Finding~1): as the policy evolves, a fixed data distribution provides diminishing corrective signal.
SAGE continues to improve steadily before converging, confirming that its refreshable pool design sustains learning by adapting to the current policy.

\begin{figure}
    \centering
    \includegraphics[width=\linewidth]{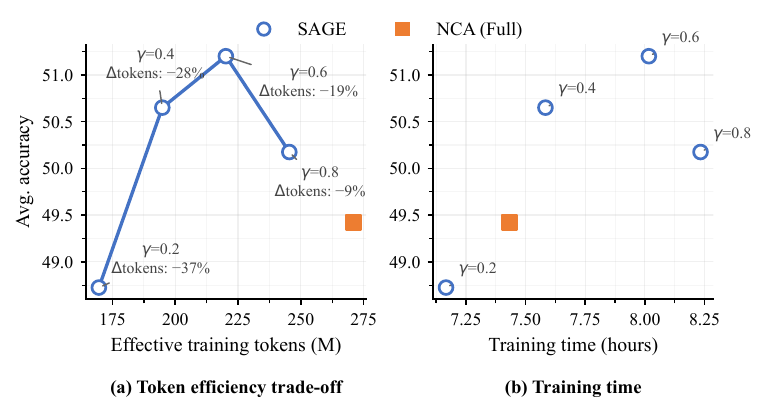}
    \caption{Accuracy versus effective training tokens (a) and wall-clock time (b) under different keep ratios $\gamma$.}
    \label{fig6}
\end{figure}

\paragraph{Data Efficiency.}
Figure~\ref{fig6}(a) shows that moderate keep ratios ($\gamma \in [0.4, 0.6]$) achieve higher accuracy than NCA (Full) while using fewer effective training tokens.
Figure~\ref{fig6}(b) confirms that wall-clock time remains comparable, since forward-only scoring is cheaper than backpropagation.

\section{Related Work}

\paragraph{Online RL for Reasoning.}
Online reinforcement learning has become a standard post-training approach for improving mathematical reasoning.
Early methods commonly used PPO~\citep{schulman2017proximal}, while recent critic-free methods such as GRPO~\citep{shao2024deepseekmath} and DAPO~\citep{yu2026dapo} learn from newly generated rollouts with reward-driven exploration.
These methods address a different stage of the post-training pipeline: they improve the policy by generating new responses online.
SAGE instead operates in the offline preference stage, where the corpus of chosen--rejected pairs is fixed, and is therefore complementary to online RL.

\paragraph{Offline Preference Optimization.}
Direct Preference Optimization (DPO)~\citep{rafailov2023direct} and its variants, including SimPO~\citep{meng2024simpo}, CPO~\citep{xu2024contrastive}, and KTO~\citep{ethayarajh2024model}, optimize the policy directly on fixed chosen-rejected pairs without online generation.
These methods focus on improving the training objective while treating all preference pairs as equally useful.
SAGE addresses a complementary dimension: given a fixed preference corpus and any pairwise objective, which pairs should receive gradient computation at each training stage.

\paragraph{Data Selection for LLM Training.}
Recent work demonstrates that data selection strongly influences training stability~\citep{zhou2024jiuzhang3,yu2024mates,li2023coltr,li2026probe}, especially for Long-CoT reasoning where optimization exhibits high gradient variance~\citep{cobbe2021training,arora2026training,zhao2026beyond}.
Static filtering and curriculum-based approaches~\citep{xia2024less,wang2024greats} improve efficiency but apply fixed criteria that cannot adapt to the evolving policy.
SAGE differs by performing policy-aware selection: it periodically re-scores candidate pairs under the current model, enabling dynamic adaptation throughout training.

\section{Conclusion}
We introduce gradient utility as a diagnostic framework for offline preference optimization in Long-CoT reasoning.
Through systematic analysis, we show that pair utility drifts with the policy, that large gradients often arise from unstable high-curvature regions, and that the most effective supervision comes from stable confident errors.
Based on these findings, we propose SAGE, which selects preference pairs by jointly considering informativeness and stability.
Experiments across multiple model scales confirm that SAGE improves reasoning accuracy and optimization stability while reducing unnecessary gradient computation.

\section*{Limitations}

SAGE operates on a static preference corpus and does not consider online data collection or exploration.
While this design keeps the method computationally efficient, performance depends on the coverage and quality of the preference data.
In addition, SAGE introduces a forward-pass scoring step for estimating sample utility, which incurs extra overhead compared to static baselines, though this cost is mitigated by the pool-based design.


\bibliography{custom}

\clearpage
\appendix
\section{Methodology Details}
\subsection{Data Construction and Filtering}\label{Data Construction}
The original Light-R1 DPO dataset\footnote{\url{https://huggingface.co/datasets/qihoo360/Light-R1-DPOData}} contains \textbf{3,000} samples. Preliminary analysis revealed substantial noise, mainly from redundant queries and \textit{degenerate preference pairs} in which the chosen and rejected responses were identical. To improve the reliability of preference learning, we applied a two-stage deduplication pipeline:

\begin{itemize}
    \item \textbf{Duplicate Query Filtering}: We removed redundant queries to reduce the risk of overfitting to specific prompts.
    \item \textbf{Preference Consistency Check}: We discarded pairs whose final answers (extracted from the \texttt{\textbackslash boxed\{\}} markers) were identical between the chosen and rejected responses.
\end{itemize}
After filtering, we obtained a high-quality subset of \textbf{1,659} samples. To restore the dataset scale, we sampled \textbf{5,000} unique instances from the SFT data\footnote{\url{https://huggingface.co/datasets/qihoo360/Light-R1-SFTData}} (as detailed in Section~\ref{section5.1}), ensuring zero overlap with the retained samples. To generate rejected responses, we used \textit{DeepScaleR-1.5B-Preview}\footnote{\url{https://huggingface.co/agentica-org/DeepScaleR-1.5B-Preview}}~\cite{luo2025deepscaler}, a small model fine-tuned specifically for reasoning tasks. By using this model, we ensured that the augmented dataset maintained the same distribution and characteristics as the original, providing consistency in the data generation process. The prompt we employed for generating rejected responses was:

\begin{quote}
    \centering \itshape
    Please reason step by step, and put your final answer within \texttt{\textbackslash boxed\{\}}.
\end{quote}

To maintain the validity of the DPO triplets, we filtered out any generated response whose final answer matched the ground-truth answer. This process produced \textbf{2,475} valid negative pairs from the 5,000 instances. Combined with the retained subset, the final dataset contains \textbf{4,134} samples.

\subsection{Preference Annotation} \label{Preference Annotation}
To implement difficulty-based partitioning for SAGE, we used \textit{DeepSeek-V3}~\cite{liu2024deepseek} as a judge to provide multi-dimensional annotations for each triplet $(x_i, y_i^w, y_i^l)$. The annotation process evaluates two dimensions to assess the optimization utility of each pair:

\begin{itemize} 
    \item \textbf{Preference Clarity}: $c_i$ measures the degree of distinctness between the chosen and rejected solutions [High, Medium, or Low].
    \item \textbf{Rejected Response Quality}: $q_i$ is an absolute quality score (1–5) assessing the logical coherence and depth of the rejected response.
\end{itemize}

We use this dual-dimension approach because the utility of a preference pair depends on the plausibility of the error. A pair exhibits \textit{High Clarity} if the rejected solution is obviously flawed, providing a strong signal for early alignment. Conversely, \textit{Low Clarity} samples, where the rejected response follows a sophisticated but flawed reasoning path, are crucial for fine-tuning the boundaries in later training stages. The judge model uses the following structured template:

\begin{quote}
\centering \small \itshape
\begin{tabular}{p{0.95\linewidth}}
\toprule
\textbf{System Prompt} \\
You are an expert mathematical evaluator. You will be presented with a math problem and two model-generated solutions: \textbf{Chosen} (correct answer) and \textbf{Rejected} (incorrect answer). \\
\\
\textbf{Task 1: Preference Clarity Assessment} \\
Evaluate the distinctness between the solutions: \\
- \textbf{High}: The rejected solution is obviously flawed; the preference is stark and easy to identify. \\
- \textbf{Medium}: The rejected solution is plausible but contains noticeable errors. \\
- \textbf{Low}: The rejected solution is highly deceptive and resembles the chosen one, requiring deep analysis to refute. \\
\\
\textbf{Task 2: Rejected Solution Quality Analysis} \\
Rate the quality of the \textbf{Rejected} reasoning path (1--5) based on its coherence and depth, despite its incorrect conclusion. \\
\\
\textbf{Format}: Return output strictly in the following JSON format: \\
\{ \\
\quad \texttt{"clarity": "High | Medium | Low",} \\
\quad \texttt{"reason": "...",} \\
\quad \texttt{"rejected\_analysis": "...",} \\
\quad \texttt{"rejected\_score": x} \\
\} \\
\midrule
\textbf{Inputs} \\
\textbf{[Problem]}: \{\{problem\_text\}\} \\
\textbf{[Chosen]}: \{\{chosen\_solution\}\} \\
\textbf{[Rejected]}: \{\{rejected\_solution\}\} \\
\bottomrule
\end{tabular}
\end{quote}

\paragraph{Analysis of Data Distribution}
The joint distribution is summarized in Table~\ref{tab:annotation_dist}. Consistent with the objectives of SAGE, most samples (3,018) fall into the \textit{High Clarity} category, providing a robust starting point for curriculum learning. The scarcity in the \textit{Low Clarity} and \textit{High Quality} regions (e.g., Q5) is expected, as highly sophisticated yet incorrect reasoning paths are rare.

\begin{table}[h]
\centering
\small
\begin{tabular}{l ccccc r}
\toprule
\textbf{Clarity} & \textbf{Q1} & \textbf{Q2} & \textbf{Q3} & \textbf{Q4} & \textbf{Q5} & \textbf{Total} \\
\midrule
High   & 946 & 1,728 & 221 & 124 & 2   & 3,018 \\
Medium        & 0   & 572   & 474 & 11  & 0   & 1,057 \\
Low    & 0   & 39    & 11  & 2   & 5   & 59    \\
\midrule
\textbf{Total} & 946 & 2,339 & 706 & 139 & 7 & \textbf{4,134} \\
\bottomrule
\end{tabular}
\caption{Joint distribution of Preference Clarity and Rejected Response Quality (Q1--Q5). High Clarity indicates obvious errors.}
\label{tab:annotation_dist}
\end{table}

\paragraph{Curriculum Subset Results}
Based on the partitioning principles in Section~\ref{section3.2}, we categorized the \textbf{4,134} triplets into three pools:
\begin{itemize}
    \item \textbf{Easy Pool} ($\mathcal{D}_{\text{easy}}$): High Clarity and $q_i \le 2$ (2,674 samples).
    \item \textbf{Medium Pool} ($\mathcal{D}_{\text{medium}}$): Medium Clarity and $q_i \in \{2, 3\}$ (1,046 samples).
    \item \textbf{Hard Pool} ($\mathcal{D}_{\text{hard}}$): Remaining Low Clarity or High Quality samples (414 samples).
\end{itemize}

\subsection{Algorithmic Details of SAGE}

For completeness, we provide the pseudocode of the SAGE training procedure in Algorithm~\ref{alg:sage}.
The algorithm summarizes the pool-based candidate refresh mechanism and the stability-aware sample selection used during training.

\begin{algorithm}[t]
\caption{SAGE: Stability-Aware Gradient Efficiency}
\label{alg:sage}
\begin{algorithmic}[1]
\STATE \textbf{Input:} Dataset $\mathcal{D}$, policy $\pi_{\theta}$, difficulty schedule $\rho(k)$, selection ratio schedule $\gamma(k)$.
\STATE Partition $\mathcal{D}$ into $\bigcup_{k=1}^{K} \mathcal{P}_k$ according to the difficulty schedule $\rho(k)$.
\FOR{$k = 1$ to $K$}
    \STATE \textbf{// On-Policy Score Evaluation}
    \STATE Perform a forward pass on $\mathcal{P}_k$ to compute length-normalized SAGE scores $\{s_i\}$ using Eq.~\ref{eq:sage_score}.
    \STATE \textbf{// Stability-Aware Filtering}
    \STATE Compute the retention count $N_k = \gamma(k) \cdot |\mathcal{P}_k|$.
    \STATE Select the high-SNR subset $\mathcal{B}_k \leftarrow \{x_i \in \mathcal{P}_k \mid \text{rank}(s_i) \le N_k\}$.
    \STATE \textbf{// Policy Optimization}
    \STATE Update the policy parameters $\theta \leftarrow \theta - \eta \nabla \mathcal{L}_{\text{NCA}}(\mathcal{B}_k)$.
\ENDFOR
\end{algorithmic}
\end{algorithm}

\section{Additional Experiments}
\subsection{Training Details}
The training hyperparameters and SAGE scheduling configurations are summarized in Table~\ref{table4}.
\begin{table}[h]
\centering
\small
\begin{tabular}{l l}
\toprule
\textbf{Hyperparameter} & \textbf{Value} \\
\midrule
Preference loss coefficient ($\beta$) & 0.1 \\
Training epochs & 3 \\
Batch size & 16 \\
Sequence parallelism (sp) & 8 \\
Maximum sequence length & 32,768 \\
Learning rate & $5\times10^{-7}$ \\
\midrule
Difficulty mix $\rho_{\text{start}}$ &
\makecell[l]{easy: 0.90 \\ medium: 0.10 \\ hard: 0.00} \\
\midrule
Difficulty mix $\rho_{\text{end}}$ &
\makecell[l]{easy: 0.40 \\ medium: 0.40 \\ hard: 0.20} \\
\midrule
Keep ratio $\gamma_{\text{start}}$ & 1.0 \\
Keep ratio $\gamma_{\text{end}}$ & 0.4 \\
\midrule
refresh step & 25 \\
\bottomrule
\end{tabular}
\caption{Training hyperparameters and SAGE scheduling configurations.}
\label{table4}
\end{table}

\paragraph{Difficulty Mixing Schedule.}
The difficulty mixing ratios are chosen empirically based on the dataset distribution after difficulty partitioning.
Specifically, the easy, medium, and hard subsets contain 2,674, 1,046, and 414 samples, respectively.
We therefore initialize training with a higher proportion of easy samples to provide stable early supervision, and gradually increase the weights of medium and hard samples through linear annealing.
This schedule balances early optimization stability with sufficient exposure to harder preference pairs in later training stages.

\subsection{Curvature Measurement via Hessian-Vector Products}
\label{appendix:curvature}
Section~\ref{sec:finding_gradient} attributes the gradient spikes of full-data training to high-curvature pairs.
Because the margin-level gradient and curvature peak at different confidence levels ($p \to 0$ and $p \approx 0.5$, respectively), the gradient-norm trace of Figure~\ref{fig3} alone cannot establish this attribution.
We therefore measure curvature directly.

For a pair $i$ with loss $\ell_i$ and gradient $u_i = \nabla_\theta \ell_i / \|\nabla_\theta \ell_i\|$, we estimate the local curvature along its own gradient direction as the Rayleigh quotient $\kappa_i = u_i^{\top} H u_i$, computed with a single Hessian-vector product (double backward on $\ell_i$), so that no explicit Hessian is formed.
We evaluate $\kappa_i$ on the Qwen2.5-1.5B-Instruct run at the checkpoints used in Section~\ref{sec:finding_dynamic} (steps 50, 150, 250), restricting the computation to the candidate pool active at each interval.
Pairs are then split by whether SAGE's curvature term discounts them: the \emph{high-curvature} group is the top quartile of $\kappa_i$ within the pool, and the \emph{low-curvature} group is the subset that SAGE retains.
A training step is counted as a \emph{spike} if its parameter-space gradient norm exceeds twice the running median over a window of 20 steps, matching the visual criterion used in Figure~\ref{fig3}.
Under this protocol, high-curvature pairs coincide with spikes in $21\%$ of the steps in which they participate, versus $5\%$ for retained low-curvature pairs.
This gap, together with the \mbox{w/o $c$} ablation in Section~\ref{section5.2}, supports the interpretation that instability is driven by curvature rather than by large margin-level gradients.

\subsection{Seed Variance and Statistical Significance}
\label{appendix:seed_variance}
The main results in Table~\ref{table1} follow the common practice of reporting a single training run per configuration.
Because several benchmarks are small and high-variance, we additionally repeat the 1.5B experiments with five random seeds (affecting data-order shuffling and pool sampling) and report mean$\pm$std in Table~\ref{table:seed_variance}.
Evaluation follows the protocol of Section~\ref{section5.1}, so AIME24 and AMC23 entries are pass@8 averages.

\begin{table}[h]
\centering
\renewcommand{\arraystretch}{1.2}
\resizebox{\linewidth}{!}{
\begin{tabular}{l ccc}
\toprule
\textbf{Benchmark} & \textbf{NCA (Full)} & \textbf{NCA (Random)} & \textbf{SAGE} \\
\midrule
GSM8K    & 74.6 $\pm$ 0.5 & 73.4 $\pm$ 0.7 & \textbf{74.9 $\pm$ 0.4} \\
MATH500  & 56.0 $\pm$ 0.8 & 55.9 $\pm$ 1.0 & \textbf{57.1 $\pm$ 0.9} \\
Minerva  & 19.3 $\pm$ 1.2 & 18.7 $\pm$ 1.5 & \textbf{20.1 $\pm$ 1.0} \\
Gaokao   & 47.5 $\pm$ 1.1 & 48.0 $\pm$ 1.4 & \textbf{50.0 $\pm$ 1.1} \\
Olymp.   & 20.2 $\pm$ 0.9 & 19.8 $\pm$ 1.1 & \textbf{21.3 $\pm$ 0.7} \\
Coll.    & 38.1 $\pm$ 0.4 & 37.9 $\pm$ 0.5 & \textbf{38.2 $\pm$ 0.5} \\
AIME24   & 9.33 $\pm$ 2.8 & 4.67 $\pm$ 3.8 & \textbf{10.0 $\pm$ 2.6} \\
AMC23    & 23.5 $\pm$ 2.9 & 24.0 $\pm$ 4.5 & \textbf{27.0 $\pm$ 3.1} \\
\midrule
\textbf{Avg.} & 36.1 $\pm$ 0.6 & 35.3 $\pm$ 0.8 & \textbf{37.3 $\pm$ 0.6} \\
\bottomrule
\end{tabular}
}
\caption{Five-seed results on Qwen2.5-1.5B-Instruct (mean $\pm$ std). Olymp.: OlympiadBench; Coll.: CollegeMath. SAGE improves the average over NCA (Full) with a paired $t$-test $p \approx 0.008$; AIME24 and AMC23 have overlapping per-seed intervals.}
\label{table:seed_variance}
\end{table}

SAGE improves the average accuracy from $36.1 \pm 0.6$ (NCA Full) to $37.3 \pm 0.6$; a paired $t$-test over the five seeds gives $p \approx 0.008$.
Per-benchmark, the gains on GSM8K, MATH500, Minerva, Gaokao, OlympiadBench, and CollegeMath are consistent in sign across seeds, whereas AIME24 and AMC23 show standard deviations of the same order as the gains themselves, with overlapping per-seed intervals.
We therefore treat the aggregate average as the primary evidence and read the competition-level numbers as directional rather than individually significant.

\subsection{Dataset Statistics}\label{Dataset}
See Table~\ref{table5} for statistics of test sets.
\begin{table}[h]
\centering
\begin{tabular}{lr}
\toprule
\textbf{Dataset} & \textbf{\# Problems} \\
\midrule
GSM8K         & 1,319 \\
MATH500       & 500   \\
Minerva  & 272   \\
Gaokao & 395   \\
Olympiad & 675   \\
CollegeMath   & 2,818 \\
AIME24 & 30 \\
AMC23 & 40 \\

\bottomrule
\end{tabular}
\caption{Statistics of the test sets used for evaluation.}
\label{table5}
\end{table}

\subsection{SFT Results}\label{sft_results}

We use the Light-R1 Stage 1 dataset to fine-tune the 1.5B-, 3B-, and 7B-Instruct models. 
The training hyperparameters follow those of Light-R1: the cutoff length is set to 20{,}000 tokens, the batch size is 128, the learning rate is $5\times10^{-6}$, and training is conducted for 10 epochs.
We evaluate checkpoints from all epochs on 8 benchmarks and select the final checkpoint based on the average accuracy across benchmarks.

\begin{figure}[t]
    \centering
    \begin{subfigure}{0.48\linewidth}
        \centering
        \includegraphics[width=\linewidth]{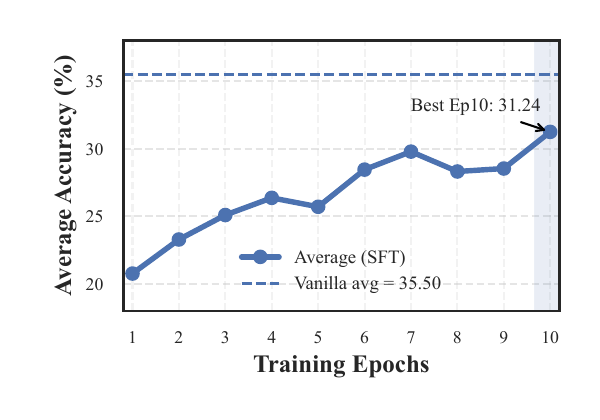}
        \caption{Qwen2.5-1.5B-Instruct}
    \end{subfigure}
    \hfill
    \begin{subfigure}{0.48\linewidth}
        \centering
        \includegraphics[width=\linewidth]{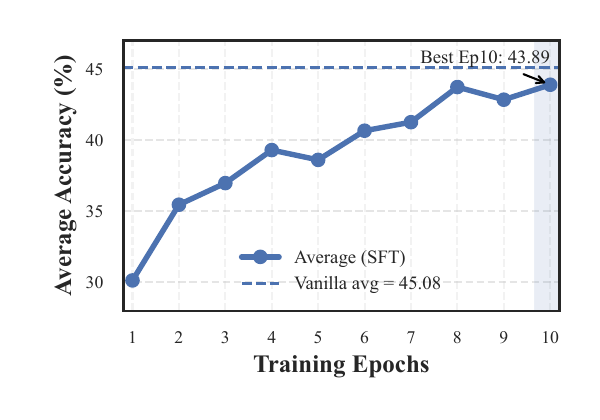}
        \caption{Qwen2.5-3B-Instruct}
    \end{subfigure}
    \caption{SFT results for small-scale models (1.5B and 3B). Although performance improves over training epochs, none of the checkpoints surpass the vanilla baselines.}
    \label{fig7}
\end{figure}

\paragraph{1.5B and 3B.}
Figure~\ref{fig7} presents the SFT results for the 1.5B and 3B models over 10 training epochs.
For both model sizes, the average accuracy increases steadily as training proceeds, indicating that supervised fine-tuning effectively improves model performance.
However, none of the SFT checkpoints within the 10 epochs surpass the corresponding vanilla baselines.
Even at their best-performing epochs, the SFT models remain below the original Qwen2.5-Instruct models in terms of average accuracy across benchmarks.

This observation suggests that, at smaller model scales, SFT alone may be insufficient to fully exploit the diverse reasoning patterns in the Light-R1 data.
Consequently, we directly adopt the original Qwen2.5-1.5B-Instruct and Qwen2.5-3B-Instruct models as the vanilla baselines for subsequent experiments.

\begin{figure}[h]
    \centering
    \includegraphics[width=\linewidth]{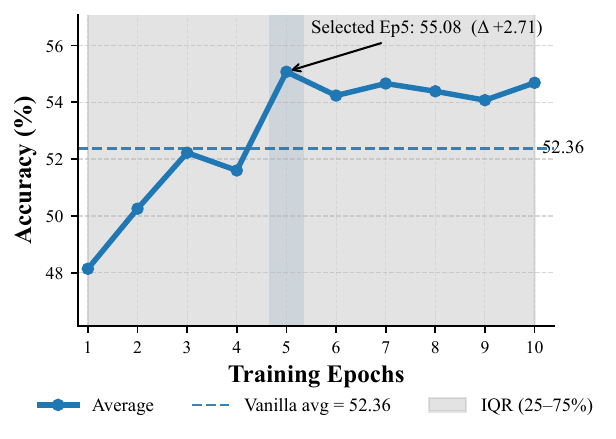}
    \caption{SFT results for the Qwen2.5-7B-Instruct model across training epochs.}
    \label{fig8}
\end{figure}

\paragraph{7B.}
The SFT results for the 7B model are shown in Figure~\ref{fig8}.
Checkpoints from epochs 5 to 10 consistently achieve higher average accuracy than the vanilla baseline, with epoch 5 yielding the best overall performance.
Accordingly, we select the epoch-5 checkpoint as the vanilla model to replace Qwen2.5-7B-Instruct in subsequent experiments.

Despite the overall improvement, the SFT model underperforms the original Qwen2.5-7B-Instruct model on \textbf{Minerva Math} and \textbf{College Math}.
Specifically, Qwen2.5-7B-Instruct achieves accuracies of 37.10 and 46.50 on these benchmarks, whereas the selected SFT checkpoint (epoch 5) attains 28.30 and 42.40, respectively.
Within the SFT training process, the best performance on Minerva Math and College Math is achieved at epoch 10 (29.40) and epoch 9 (43.20), respectively, indicating that these benchmarks may benefit from later-stage fine-tuning.

\subsection{Sensitivity to Pool Refresh Frequency}
\begin{figure}
    \centering
    \includegraphics[width=0.9\linewidth]{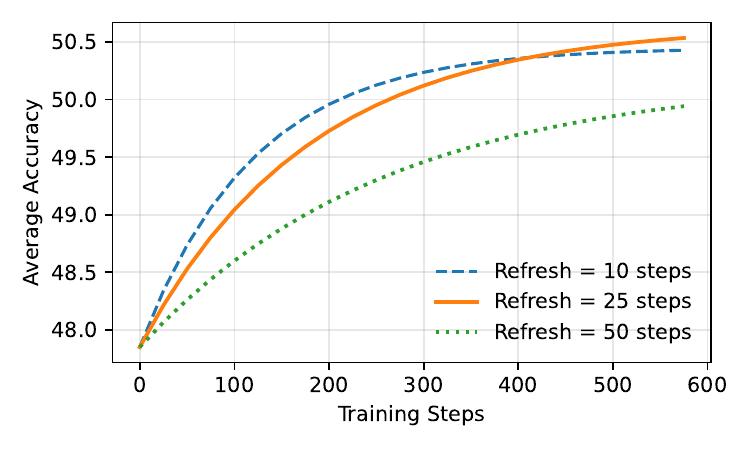}
    \caption{Sensitivity of SAGE to candidate-pool refresh frequency.}
    \label{fig9}
\end{figure}

We analyze the sensitivity of SAGE to the candidate-pool refresh frequency by varying the refresh interval while keeping all other hyperparameters fixed.
As shown in Figure~\ref{fig9}, all configurations start from the same initial performance and improve consistently during training, indicating that SAGE is not sensitive to a specific refresh schedule.
A moderate refresh interval (25 steps) achieves the best final performance, while both more frequent and sparser refreshing lead to slightly lower accuracy.
This trend reflects a trade-off intrinsic to SAGE: overly sparse refreshing yields stale utility estimates under a changing policy, whereas overly frequent refreshing provides diminishing returns while increasing scoring overhead.
Overall, SAGE exhibits a broad and stable operating region with respect to the pool refresh frequency.

\subsection{Validation of LLM-Based Difficulty Annotation}
\label{appendix:judge_validation}
The difficulty annotations used in SAGE are generated by an LLM-as-judge (DeepSeek-V3).
We address potential concerns regarding reliability, cost, and the relationship to knowledge distillation.

\paragraph{Annotation Reliability.}
We manually verified a stratified subset: all 59 Low-Clarity samples, 20\% of Medium-Clarity samples (from 1,057), and 10\% of High-Clarity samples (from 3,018).
The agreement rate between LLM annotations and human verification is 96\%, indicating strong label consistency.
The high agreement is expected because the annotation task (assessing whether a rejected response is obviously flawed vs.\ subtly wrong) is well-defined and does not require subjective judgment.

\paragraph{Computational Cost.}
The annotation is performed \emph{once} during data preprocessing and incurs no overhead during training.
Using the DeepSeek-V3 API, annotating all 4,134 samples costs approximately \$3 USD and completes in under 20 minutes with batched requests.
This is negligible compared to the GPU cost of preference training (8$\times$A800 for multiple hours) and can be amortized across arbitrary numbers of training runs.

\paragraph{Distinction from Knowledge Distillation.}
The LLM judge provides only coarse categorical labels (High/Medium/Low clarity and a 1--5 quality score) for \emph{pool construction}, not for training signal.
This is fundamentally different from knowledge distillation, which transfers the teacher's output distribution or reasoning traces into the student's parameters.
SAGE's training signal comes entirely from the preference pairs themselves via the NCA loss; the judge annotations merely determine which pool a sample enters.
Replacing the LLM judge with any other difficulty estimator (e.g., human annotation, heuristic rules based on response length or error type) would yield a functionally equivalent system, confirming that the judge is not a source of distilled knowledge.

\subsection{Relationship to Online RL Methods}
\label{appendix:grpo_discussion}
A natural question is whether SAGE should be compared directly with online RL methods such as GRPO~\citep{shao2024deepseekmath} or DAPO~\citep{yu2026dapo}.
We argue that such a comparison conflates two distinct stages of the post-training pipeline.

\paragraph{Pipeline Stage Distinction.}
Modern reasoning model training typically follows a multi-stage pipeline (e.g., Light-R1~\citep{wen2025light}): (1) supervised fine-tuning on curated reasoning traces, (2) offline preference optimization on fixed chosen-rejected pairs, and optionally (3) online RL with reward-driven exploration.
GRPO operates in Stage~3: it generates new responses via rollouts, evaluates them with a reward signal, and updates the policy through exploration.
SAGE operates in Stage~2: it selects which \emph{existing} preference pairs should contribute gradients at each training step.
These stages address fundamentally different questions (``how to explore'' vs.\ ``how to allocate supervision'') and are complementary rather than competing.

\paragraph{Composability.}
SAGE and online RL can be composed sequentially in a pipeline: one can first apply SAGE for efficient offline preference alignment, then follow with GRPO for further exploration-based improvement.
The two methods are not substitutes; comparing them directly would be analogous to comparing a data augmentation strategy with a model architecture, both of which can be applied independently and jointly.
\label{sec:appendix}

\end{document}